\documentclass[letterpaper]{article} % DO NOT CHANGE THIS
\usepackage{aaai24}  % DO NOT CHANGE THIS
\usepackage{times}  % DO NOT CHANGE THIS
\usepackage{helvet}  % DO NOT CHANGE THIS
\usepackage{courier}  % DO NOT CHANGE THIS
\usepackage[hyphens]{url}  % DO NOT CHANGE THIS
\usepackage{graphicx} % DO NOT CHANGE THIS
\usepackage{natbib}  % DO NOT CHANGE THIS AND DO NOT ADD ANY OPTIONS TO IT
\usepackage{caption} % DO NOT CHANGE THIS AND DO NOT ADD ANY OPTIONS TO IT
\usepackage{algorithm}
\usepackage{algorithmic}

\usepackage{newfloat}
\usepackage{listings}
\DeclareCaptionStyle{ruled}{labelfont=normalfont,labelsep=colon,strut=off} % DO NOT CHANGE THIS
\floatstyle{ruled}
\newfloat{listing}{tb}{lst}{}
\floatname{listing}{Listing}
\title{BLIP-Adapter: Parameter-Efficient Transfer Learning for Mobile Screenshot Captioning}
\author{
    Ching-Yu Chiang, I-Hua Chang, Shih-Wei Liao
}
\affiliations{
    National Taiwan University\\
    r10922141@csie.ntu.edu.tw, r11922136@csie.ntu.edu.tw, liao@csie.ntu.edu.tw
}

\usepackage{bibentry}
\begin{document}

\maketitle

\begin{abstract}
This study aims to explore efficient tuning methods for the screenshot captioning task.
Recently, image captioning has seen significant advancements, but research in captioning tasks for mobile screens remains relatively scarce.
Current datasets and use cases describing user behaviors within product screenshots are notably limited.
Consequently, we sought to fine-tune pre-existing models for the screenshot captioning task.
However, fine-tuning large pre-trained models can be resource-intensive, requiring considerable time, computational power, and storage due to the vast number of parameters in image captioning models.
To tackle this challenge, this study proposes a combination of adapter methods, which necessitates tuning only the additional modules on the model.
These methods are originally designed for vision or language tasks, and our intention is to apply them to address similar challenges in screenshot captioning.
By freezing the parameters of the image caption models and training only the weights associated with the methods, performance comparable to fine-tuning the entire model can be achieved, while significantly reducing the number of parameters.
This study represents the first comprehensive investigation into the effectiveness of combining adapters within the context of the screenshot captioning task.
Through our experiments and analyses, this study aims to provide valuable insights into the application of adapters in vision-language models and contribute to the development of efficient tuning techniques for the screenshot captioning task.
Our study is available at https://github.com/RainYuGG/BLIP-Adapter
\end{abstract}

\section{Introduction}

Recently, in this era where everyone owns a smartphone, screenshot captioning has attracted increasing attention.
This task is aimed at producing natural language descriptions of user behaviors captured within mobile screenshots.
Without these screenshot captioning systems, users are burdened with the task of manually describing the UI of mobile applications whenever they need to report issues to developers, or when creating application tutorials, etc.
This process can be both time-consuming and labor-intensive.
Our objective is to investigate efficient tuning strategies tailored for the screenshot captioning task.

Machine learning has achieved significant success in both vision and language tasks \cite{dosovitskiy2021an, devlin-etal-2019-bert, raffel2020exploring, liu2021swin, yang2022moat, NIPS2017_3f5ee243}.
Moreover, there have been notable advancements in vision-language tasks \cite{radford2021learning, li2022lavis, li2021align, li2022blip, li2023blip, DBLP:journals/corr/abs-1909-11059, zhou2022learning}, such as image-text matching, visual question answering, and image captioning. In these frameworks, Vision-language models, which typically utilize both a visual model and a language model, have greatly improved due to enhanced architecture designs and the availability of large-scale high-quality datasets \cite{10.1007/978-3-319-10602-1_48, 5206848, krishna2017visual, xu2016msr, chen-dolan-2011-collecting, agrawal2019nocaps}.
Despite advancements in architectural designs, the size of modern vision-language models is rapidly increasing, leading to substantial memory and storage requirements. Moreover, these models typically comprise a vast number of parameters, which poses a significant challenge for training from scratch and consumes considerable time and computational resources.
Large-scale high-quality datasets in image captioning tasks primarily consist of general real-world scenes, with a lack of labeled datasets specifically tailored to special domains, such as medical images \cite{subramanian2020medicat}, earth observation images \cite{ZIA2022102741}, and mobile screenshots \cite{wang2021screen2words}.
However, collecting large-scale datasets for every visual task is labor-intensive and prohibitively expensive to scale. Moreover, ensuring high-quality datasets also requires human resources for data labeling.
To overcome these problems, fine-tuning the pre-trained models has been widely adopted modernly.
The models pre-trained on large-scale datasets, such as ImageNet \cite{5206848}, COCO \cite{10.1007/978-3-319-10602-1_48}, and Visual Genome \cite{krishna2017visual}, are used.
These models are then subsequently fine-tuned on smaller-scale specific downstream datasets.
It serves as a solution to leverage the pre-training knowledge from large-scale datasets and adapt it to specific tasks at hand.

\begin{figure}
    % \centerline{\includegraphics[height=8.4cm, width=12.8cm]{figures/fig1.png}}
    \centerline{\includegraphics[width=\linewidth]{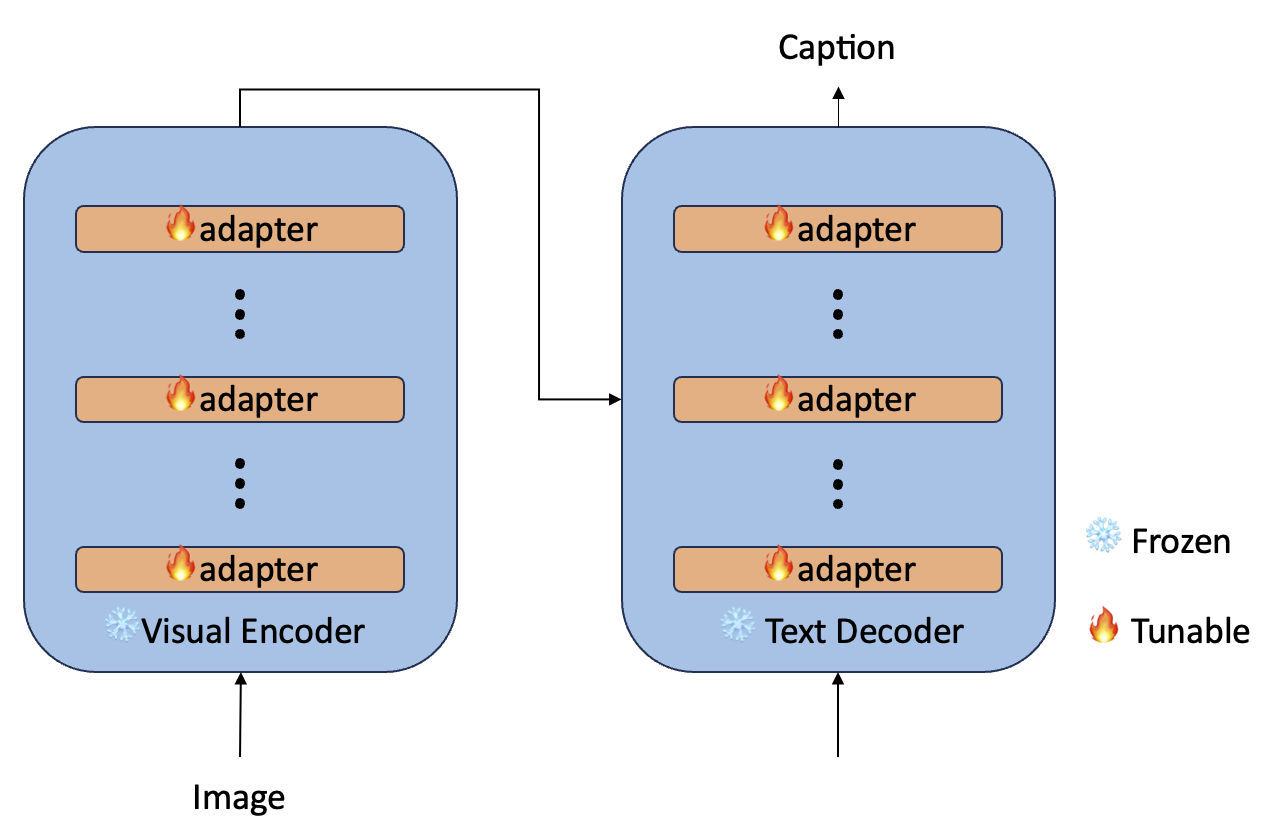}}
    \caption{
        Illustration of the adapter-based fine-tuning approach.
        Through the insertion of adapters, we can selectively fine-tune the lightweight components, which, in our case, comprise approximately 1.47\% of the model's total parameters.}
    \label{fig:1}
\end{figure}

\setlength{\parindent}{4.5ex}
However, fine-tuning the entire model can still be resource-intensive, consuming considerable computation power, storage, and time, especially for large vision-language models.
To address this, adapter-based fine-tuning \cite{houlsby2019parameter, guo2020parameter, chen2023sam, sung2022vl} has emerged as a more parameter-efficient alternative.
Similarly, the model pre-trained on large-scale datasets are used as the backbones and the adapters with light learnable weight will be inserted into the model.
As shown in Figure \ref{fig:1}, by using adapters, the weights in the pre-trained models are frozen, subsequently, fine-tuning the adapters on smaller-scale downstream datasets.
we can only conduct training on additional adapters to get the fine-tuning effect. 
This approach enables the model to be adapted to specific tasks while preserving the knowledge acquired during pre-training.
Additionally, it helps mitigate the limitations related to dataset availability, computational resources, and storage requirements that are often encountered when training models from scratch.

%  The adapters have been successfully applied in various domains, including natural language processing, computer vision, and multimodal learning \cite{houlsby2019parameter, DBLP:journals/corr/abs-2110-04366, sung2022vl, sung2022vladapt}
% As shown in figure~\ref{fig:1}, we can only tune the lightweight adapters to achieve the fine-tuning effect, which is more efficient than tuning the entire model.
% Adapters have facilitated advancements in various machine learning tasks by effectively transferring knowledge from pre-training to task-specific adaptation.
\setlength{\parindent}{4.5ex}
To address the screenshot captioning task, this study explores various methods and techniques for implementing parameter-efficient tuning and evaluates their effectiveness within this specific context.
Modifications of the state-of-the-art vision-language model, BLIP, are explored by employing parameter-efficient tuning methods for task-specific fine-tuning in the mobile user interface (UI) screenshot captioning domain.
The impacts of various methods are assessed, providing a comprehensive evaluation of their respective effects.
Additionally, a modification is incorporated into the BLIP model, similar to the implementation in previous work, by inserting an intermediate layer between the vision and language models.
The effect of this alteration, coupled with the use of parameter-efficient methods on the language model, is analyzed.
Furthermore, this study experiments with different combinations of parameter-efficient tuning strategies on BLIP, evaluating their efficacy for screenshot captioning tasks.
Our contributions could be summarized as: 
\begin{itemize}
    \item The evaluation of various parameter-efficient tuning strategies is conducted, applied separately to vision and language tasks, on the state-of-the-art captioning model, BLIP.
    % \item We conduct an evaluation of various parameter-efficient tuning strategies, which are applied separately to vision and language tasks, on the state-of-the-art captioning model, BLIP.
    \item This study presents comprehensive transfer learning research applied to Screen2Words, a dataset specifically tailored for image captioning tasks within the mobile UI domain.
    % \item We present the comprehensive transfer learning research applied to Screen2Words, a dataset specifically tailored for image captioning tasks within the mobile UI domain.
    \item The demonstration that applying a combination of different parameter-efficient tuning methods can achieve performance comparable to full fine-tuning, but only requires updating 0.08\% to 1.47\% of the parameters.
    % \item We demonstrate that, by applying the combination of the different parameter-efficient tuning methods, we can achieve performance comparable to full fine-tuning, while only needing to update 0.08\% to 1.47\% of the parameters.
\end{itemize}

\section{Related Work}
\subsection{Vision-Language Models}
Vision-language models are a category of models that blend vision and language components to tackle tasks that cross both domains. These models' architectures may vary depending on the specific task.
For instance, in image-text matching tasks, a Siamese network architecture, featuring both a visual encoder and a text encoder, is often preferred.
Conversely, image captioning tasks typically employ an encoder-decoder architecture, which comprises a visual encoder and a text decoder.

Over the years, a variety of vision-language models have emerged, reflecting advancements in architectural designs and pre-training strategies, especially in image captioning tasks.
In 2015, Vinyals et al. \cite{vinyals2015show} introduced an image caption generator that synergizes Convolutional Neural Networks \cite{krizhevsky2012imagenet} as a visual encoder with Recurrent Neural Networks as a text decoder.
% This model undergoes pre-training on a large-scale dataset before it is fine-tuned for image captioning.
In 2018, Anderson et al. introduced the Bottom-Up and Top-Down Attention model \cite{anderson2018bottom}.
Its primary innovation lies in the utilization of Faster R-CNN \cite{ren2015faster} for object detection, enabling the achievement of bottom-up attention by obtaining the corresponding detected targets and labels.
Further, Long-Short Term Memory (LSTM) networks \cite{hochreiter1997long} were leveraged in the decoder, dynamically adjusting the focus on the input image features according to the output language.
This attention mechanism allows the model to pay greater attention to the more salient and important objects in the image, thus creating a better description.
% In 2018, Anderson et al. proposed the Bottom-Up and Top-Down model \cite{anderson2018bottom}. This model employs a Faster R-CNN network \cite{ren2015faster} in the visual encoder to get the objects and generate region-level features, and pairs it with a Long-Short Term Memory (LSTM) network \cite{hochreiter1997long} serving as the text decoder.
% This model achieves state-of-the-art results by leveraging the rich contextual information present in both the global and local image regions.
However, after transformer-architecture-based models \cite{NIPS2017_3f5ee243, devlin-etal-2019-bert} demonstrated state-of-the-art performance in natural language processing tasks, their impact extended to the field of vision-language models.
The introduction of the Vision Transformer by Dosovitskiy et al. in 2020 \cite{dosovitskiy2021an} marked a significant milestone.
The Vision Transformer applies the transformer architecture to visual data by treating images as sequences of patches.
This approach achieved competitive results in image classification tasks and influenced the development of vision-language models.
Building on these innovations, the BLIP model, introduced by Li et al. in 2022 \cite{li2022blip}, has achieved state-of-the-art performance across multiple vision-language tasks.
It achieves this through pre-training multimodal components such as visual and text encoders, as well as a text decoder, which are applicable to various vision-language tasks, including image-text matching, visual question answering, and image captioning.
During pre-training, BLIP utilizes large-scale datasets, including COCO \cite{10.1007/978-3-319-10602-1_48}, Visual Genome \cite{krishna2017visual}, Conceptual Captions \cite{sharma2018conceptual}, Conceptual 12M \cite{changpinyo2021conceptual}, SBU captions \cite{ordonez2011im2text}, and LAION \cite{schuhmann2021laion} for comprehensive pre-training.
In image captioning tasks, the visual encoder and text decoder components are used, which are ViT \cite{dosovitskiy2021an} and BERT \cite{devlin-etal-2019-bert} respectively, followed by fine-tuning on the COCO Caption dataset \cite{10.1007/978-3-319-10602-1_48}.
This convergence of ViT and BERT underlines the transformative role of the transformer architecture in propelling advancements in vision-language models.
Such progress highlights the importance of architectural innovation and refined pre-training strategies in further advancing the field of vision-language understanding and generation.

\subsection{Adapter Approaches}
Adapters, introduced by Houlsby et al. in 2019 \cite{houlsby2019parameter}, are a parameter-efficient tuning technique employed in transfer learning.
This approach involves adding lightweight, task-specific layers or modules to a pre-trained model without altering the original parameters.
Instead of fine-tuning the entire model, it's feasible to tune only the small-scale adapters to achieve the fine-tuning effect.
By selectively updating these adapters while freezing the remaining parameters, parameter efficiency is achieved without significantly compromising the model's performance. 
Moreover, by only saving the additional weights of the adapters during training, greater storage efficiency is achieved compared to saving the weights of the entire model.
These adapters learn task-specific information while keeping the pre-trained model's parameters intact, thus reducing computational and storage costs and allowing for better generalization to different tasks.

Adapters have shown significant performance not only in natural language processing tasks \cite{houlsby2019parameter, DBLP:journals/corr/abs-2110-04366, bapna2019simple, pfeiffer2020AdapterHub, chen2023exploring} but also in various vision tasks. \cite{chen2023sam, pan2022st, Ermis_2022_CVPR, sung2022vladapter}.
The bottleneck adapters, as proposed by Houlsby et al., are inserted into the transformer architecture, specifically after the feed-forward layers. These adapters comprise a down projection, followed by a GELU activation function, and then an up projection. 
Bapna et al. \cite{bapna2019simple} modify the adapter architecture for translation tasks by incorporating an additional layer normalization and replacing the GELU activation function with a ReLU activation function.
% Compacter
The Compacter architecture, proposed by Mahabadi et al. in 2021 \cite{karimi2021compacter}, modifies the adapter structure by replacing the linear down- and up-projections with a parameterized hypercomplex multiplication layer. Distinct from the linear layer, this hypercomplex multiplication layer generates its weight matrix from two smaller matrices, consequently decreasing the total number of parameters. Additionally, these matrices can be factorized and shared across all adapter layers.
% Prefix Tuning
Prefix Tuning, introduced by Li et al. in 2021 \cite{li2021prefix}, innovates by incorporating new parameters within the multi-head attention blocks in each transformer layer. Specifically, it enhances the model by prepending trainable prefix vectors to the keys and values of the attention head input, which adds flexibility and adaptability to the model's attention mechanism.
% LoRA
Low-Rank Adaptation (LoRA), introduced by Hu et al. in 2021 \cite{hu2021lora}, incorporates trainable low-rank decomposition matrices into the layers of a pre-trained model.
Specifically, LoRA targets the attention weights within the transformer's self-attention sub-layers.
% BitFit
Instead of introducing additional parameters, BitFit \cite{zaken2021bitfit} simply fine-tunes the bias terms within each module, allowing for task-specific adaptation with minimal changes to the pre-trained model.
% EVP
In object detection tasks, Liu et al. introduced Explicit Visual Prompting (EVP) in 2023 \cite{liu2023explicit}, which employs an architecture similar to Houlsby's adapters.
EVP distinguishes itself by incorporating handcrafted features as input to the adapters and utilizing shared up-projection layers within the adapter structure.
% EXPLORING EFFICIENT-TUNING METHODS IN SELF-SUPERVISED SPEECH MODELS
% VL-Adapter
In 2022, Sung et al. introduced VL-Adapter \cite{sung2022vladapter}, wherein they experimented with integrating various adapters into vision-language models and evaluated their performance. They primarily applied these adapters to video question answering tasks. In contrast to our approach, their model features both a visual encoder and a text encoder, as well as a text decoder, while ours does not include a text encoder. Additionally, their models are pre-trained on general image datasets and NLP tasks, whereas our model is specifically pre-trained for the image captioning task and we further fine-tuned for a specialized screenshot captioning dataset.

\subsection{Mobile Screenshot Captioning}
Mobile screenshot captioning is a specialized subset of image captioning that concentrates on generating textual descriptions for mobile screenshots.
This task is notably challenging due to the distinctive traits of mobile screenshots, including the presence of various UI elements, the absence of fixed layouts, and a wide range of UI elements and styles.
Contrary to general image captioning, which emphasizes describing the overall scene and objects, mobile screenshot captioning focuses on articulating the functionality and content of the UI elements.
Additionally, in mobile screenshot captioning, the layout of UI elements is more closely linked to captioning performance.

% There are several previous works that have contributed to mobile screenshot captioning.
Screen2Words, introduced by Wang et al. in 2021 \cite{wang2021screen2words}, is the first open-source dataset for mobile screenshot captioning.
This dataset is built upon the Rico dataset \cite{deka2017rico}, which is a large-scale dataset containing mobile app interface images, and UI layouts.
Screen2Words enhances the mobile screenshots in the Rico dataset \cite{deka2017rico} by including human-labeled textual descriptions that correspond to the screenshots.
The dataset consists of 22,417 unique Android screenshots, each accompanied by five concise language descriptions that convey important content and functionality of the mobile screens.
These descriptions are valuable for various language-based application scenarios.
Additionally, they offer a model that includes a ResNet encoder and Transformer decoder, which serves to evaluate the performance of the dataset.

\section{Methodology}
\label{ch:3}
We explored the effectiveness of various parameter-efficient tuning strategies as applied to screenshot captioning tasks.
Our goal is to illustrate how employing a combination of parameter-efficient tuning methods can contribute to achieving parameter-efficient tuning for our caption models while maximizing the performance of screenshot captioning systems.
The BLIP Caption model \cite{li2022blip} is employed as our image caption model and is fine-tuned by the Screen2Words dataset \cite{wang2021screen2words} as our baseline.
Subsequently, the following parameter-efficient tuning strategies were evaluated separately on the visual encoder and text decoder of the BLIP Caption model.
Following this, different combinations of these strategies were employed to ascertain the most effective approach for optimizing the model architecture and fine-tuning the components of the model to achieve desired outcomes.
The specifics of these parameter-efficient tuning approaches and the model architecture are detailed in the following sections.

\subsection{Parameter-Efficient Tuning approaches}

As our model is a vision-language model, we had explored the usage of adapters specifically designed for the vision and language components separately.
These adapters, including the Houlsby adapter \cite{houlsby2019parameter}, BitFit \cite{zaken2021bitfit}, LoRA \cite{hu2021lora}, and Explicit Visual Prompting \cite{liu2023explicit}, were examined both individually and in various combinations in the experiments.

\begin{figure}
    % \centerline{\includegraphics[height=11.25cm, width=20cm]{figures/fig2.png}}
    \centerline{\includegraphics[width=\linewidth]{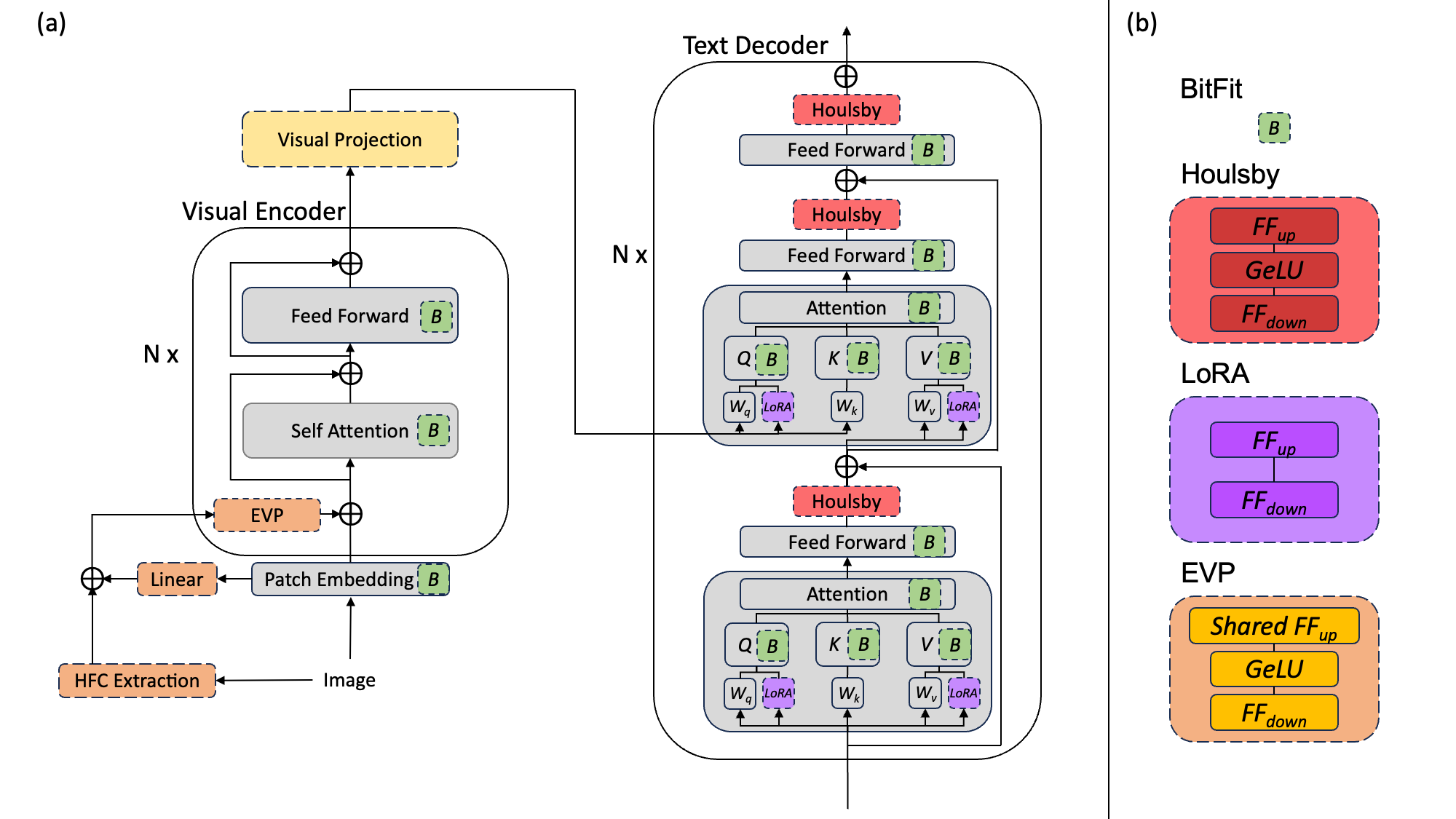}}
    \caption{
        Illustration of the adapter insertion points in the BLIP Caption model.
        (a) The modifications to the BLIP Caption model architecture.
        (b) The architecture of parameter-efficient method modules.
    }
    \label{fig:2}
\end{figure}

\begin{itemize}
    \item \textbf{Houlsby adapter} is additional bottleneck module that consist of a down-projection, GELU activation, an up-projection, and a skip connection.
In our implementation, Houlsby adapters were integrated into the three feed-forward layers of each transformer block in the text decoder as shown in Figure \ref{fig:2} (a).
    \item \textbf{BitFit} stands out by not necessitating the introduction of additional modules to the model.
Instead, it focuses on fine-tuning the biases of the existing modules within the model.
In these experiments, the effectiveness of applying BitFit to either the visual encoder or the text decoder is examined.
    \item \textbf{LoRA} consists of a down-projection and an up-projection that run in parallel with the existing linear-projection layers.
In our experiments, LoRA was integrated into the attention modules of the transformer blocks in text decoder as shown in Figure \ref{fig:2} (b).
The attention module typically consists of queries, keys, and values, with LoRA being employed in the queries and values.
    \item \textbf{Explicit Visual Prompting} (EVP) adapters utilize a similar architecture to Houlsby adapter, comprising a down-projection, GELU activation, and an up-projection.
However, there are notable differences in terms of input, weight sharing, and network connection.
In the case of EVP adapters, the input includes both the projection of the input image and task-specific information.
Furthermore, the weight of the up-projection is shared among all EVP adapters, allowing for efficient parameter sharing.
These adapters are inserted in front of each transformer block within the visual encoder.
\end{itemize}

\subsection{Model Architecture Modification}

As we need to insert additional modules into the model and freeze the original weights except for those we want to fine-tune, there are several ways to modify the model architecture.

First, experiments were conducted using these parameter-efficient tuning methods individually, either on the visual encoder or the text decoder, to observe the impact of each method on the vision-language model.
Since EVP adapter is specifically tailored for object detection tasks and aligns well with the Vision Transformer architecture, it was deployed on the visual encoder.
To this end, three different handcrafted features were experimented with as inputs to EVP adapter: the Fast Fourier Transform of the original image, as outlined in EVP paper, the original image itself, and its grayscale version.
The inclusion of the grayscale version is particularly beneficial for screenshots, as the semantics of most screenshot elements are color-irrelated, which is described in the Screen2Words paper.
On the other hand, given that the other methods are primarily engineered for NLP tasks, it was considered fitting to apply them to the text decoder.
This was done to determine whether the method modified on the visual encoder or the text decoder is more effective for the vision-language model when choosing only one of them.
The positions of insertions or modifications for each method within the model are depicted in Figure \ref{fig:2} (a).

Second, a modification similar to the model implementation of the VL-adapter \cite{sung2022vl} was implemented, which involves inserting a visual projection between the visual encoder and the language model as shown in Figure \ref{fig:2} (a).
A key difference between our model and that of the VL-adapter is the absence of a text encoder in the former.
Therefore, the linear projection layer was inserted between our visual encoder and text decoder.
In this strategy, only the visual projection and associated modules of the methods on the text decoder were fine-tuned.
Given the freeze applied to the entire visual encoder, only the methods on the text decoder were employed.
Moreover, an experiment was conducted using the Vision Transformer block as the visual projection instead of the linear projection layer, inspired by the implementation of the VL-adapter.
This was done to determine whether employing a visual projection and solely fine-tuning the text decoder could outperform fine-tuning the entire model in this particular scenario.

Finally, an attempt was made to combine the methods on both the visual encoder and text decoder to see if the combination could lead to enhanced results.
EVP adapters were integrated into the visual encoder using various handcrafted features, and combined with Houlsby adapter and LoRA, which were implemented on the text decoder.
As the underlying principle of BitFit is not to introduce additional modules but rather to fine-tune the biases, it was additionally tested across the entire model.

Through these experiments, insights were gained into how different adapters and methods affect the vision-language model's performance.
Our findings provide guidance for selecting the optimal approach in fine-tuning and architecting the model to achieve the desired results in screenshot captioning systems.

\section{Evaluation}
We begin by introducing the experimental settings, including dataset selection, evaluation metrics, and training procedures.
Subsequent utilization of parameter-efficient tuning methods and evaluation of various model structure modifications associated with these methods was carried out, as detailed in the Methodology section.
Initially, the performance of the methods when applied to the visual and language components was evaluated individually.
Thereafter, the visual encoder was frozen and only the text decoder and the inclusion of a visual projection were trained.
Finally, the integration of the methods into both the visual and language components of the BLIP caption model was explored.
In this three attempts, the entire model with approximately 223 million parameters, which underwent complete fine-tuning, served as a baseline for performance comparisons.
This enables the assessment of how closely each modification and the methods can approach the baseline performance.
Through these evaluations, the aim was to determine the impact of each modification and the methods on the performance and parameter efficiency of the model.
This comprehensive analysis enabled the identification of the most effective strategies for enhancing the screenshot captioning system.

\subsection{Experiment Settings}
In our experiments, the Screen2Words dataset \cite{wang2021screen2words} is utilized for training, validation, and evaluation.
The dataset was split according to the guidelines provided in its release.
Each screenshot in the dataset is accompanied by five captions. 
To match the number of captions and ensure sufficient training data, the screenshots were duplicated during the training phase, resulting in five training instances for each.
During the evaluation phase, the five captions associated with each screenshot were used as references.
To gauge the performance of the models, the BLEU-4 and CIDEr metrics \cite{papineni2002bleu, vedantam2015cider} were employed, both being widely used benchmarks in the field of image captioning.
All experiments were conducted using the PyTorch deep learning framework \cite{paszke2019pytorch} and were performed on a single Nvidia RTX A6000 GPU.
During the training process, the AdamW optimizer with a weight decay of 0.1 was utilized, and the learning rate was set to 5e-5.
The batch size for training is 32, and the models were trained for a total of 30 epochs.

\subsection{Individual Tuning of Visual and Language Components}
The parameter-efficient tuning approaches were employed for the visual encoder and text decoder separately to observe the individual effects of each approach on our captioning task.
For the text decoder, methods such as BitFit, Houlsby adapter, and LoRA were utilized.
In this scenario, the entire visual encoder was frozen and the training was focused on the components of the method integrated into the text decoder.
Conversely, for the visual encoder, methods such as EVP, BitFit were employed, and the entire text decoder was frozen while training exclusively focused on the components associated with the method on the visual encoder.
Additionally, due to the use of handcrafted feature extraction in EVP, three different handcrafted features were employed: the Fast Fourier Transform of the original image, along with the original image and its grayscale version.
In our notation, ``FT" stands for fine-tuning, ``EVP" represents Explicit Visual Prompting, ``EVP-gs" indicates Explicit Visual Prompting with grayscale extraction, and ``EVP-fft" signifies Explicit Visual Prompting with Fast Fourier Transform extraction.
Additionally, ``(A)" denotes the entire model, ``(V)" corresponds to the visual encoder, and ``(T)" signifies the text decoder.
The ``Parameters(\%)" column in the table denotes the percentage of trainable parameters in the model.
The results are shown in Table \ref{tab:1}.

% tell the result
We can observe that by fine-tuning the entirety of a model component if it has a sufficient number of parameters, the performance can either approach the baseline or even show potential for enhancement.
Both fine-tuning the entire visual encoder and the entire text decoder could achieve more than 90\% BLEU-4 score of the baseline.
Notably, when solely fine-tuning the entire visual encoder, the performance exceeds expectations, with the CIDEr score surpassing that achieved by fine-tuning the whole model.
We guess that this can be attributed to the text decoder, which remains frozen, already having been pre-trained for the captioning task, thus proving sufficient in generating quality captions.
Additionally, the sufficient parameters of the visual encoder are conducive to the learning of task-specific information.
% adapter's result
When employing parameter-efficient tuning methods individually, Houlsby adapter stands out by achieving the highest performance, reaching approximately 91.4\% of the CIDEr score attained by fine-tuning the entire model, thereby surpassing all other methods.
Additionally, a marked decline in performance was observed when these methods were utilized in isolation, especially if they were not applied to the text decoder.
This can be attributed to the fewer parameters being adjusted, which suggests that methods applied solely to the visual encoder may not possess sufficient parameters to learn task-specific information, whereas methods applied exclusively to the text decoder are more adept at learning how to generate captions.
Therefore, it is more effective to use the methods on the text decoder and fine-tune it to achieve satisfactory results.
This enables the model to adjust and enhance its language generation capabilities for the specific task, leading to getting close to the performance achieved by fine-tuning the entire model.

\hfill \break

\renewcommand{\arraystretch}{1.5}

\begin{table}[t]
    \centering
    \resizebox{.95\columnwidth}{!}{
    % \setlength{\tabcolsep}{12pt}
    % \begin{tabular}{p{4cm}|p{3cm}p{3cm}p{3cm}}
    \begin{tabular}{c|ccc}
    \hline
    & BLEU-4    & CIDEr & Parameters(\%) \\ \hline                           
    FT (A)       & 20.6 & 88.3 & 100.0 \\
    FT (V)     & 18.6 & \textbf{89.6} & 38.44 \\
    FT (T)     & \textbf{19.0} & 82.3 & 61.56 \\
    \hline
    EVP (V)             & 11.6 & 65.1 & 0.29 \\
    EVP-gs (V)          & 11.7 & 64.9 & 0.29 \\
    EVP-fft (V)         & 12.1 & 66.1 & 0.29 \\
    BitFit (V)          & 13.4 & 67.9 & \textbf{0.05} \\
    BitFit (T)          & 14.8 & 70.4 & 0.08 \\
    Houlsby (T)         & \textbf{18.0} & \textbf{80.7} & 1.18 \\
    LoRA (T)            & 14.5 & 74.2 & 0.26 \\
    \hline
    \end{tabular}
    }
    \caption{Performance of individual component tuning using parameter-efficient methods.}
\label{tab:1}
\end{table}

\subsection{Text Decoder and Visual Projection Tuning}
The visual encoder was frozen and focus was exclusively put on training the text decoder and the visual projection layer as illustrated in Figure \ref{fig:2} (a).
BitFit, Houlsby adapter, and LoRA are utilized with a visual projection layer to investigate the potential performance enhancements this combination might yield.
The visual projection layer employed is a linear projection layer that maps the visual features to the same dimensionality as the original features.
Moreover, an experiment was conducted using a Vision Transformer block (ViT block) as an alternative to the linear projection layer.
The results are shown in Table \ref{tab:2}.

% explain the result
We observe that when using a visual projection layer and training both the projection and the entire text decoder, the performance significantly declines, with the CIDEr score dropping to 67.3, which is approximately 76.2\% of the baseline.
This decline can be attributed to the visual projection layer being initialized with zeros, which destabilizes the pre-trained weights on the text decoder during training.
In contrast, due to the frozen pre-trained weights on the text decoder, LoRA and Houlsby adapter only fine-tune the additional parameters, making them more stable.
In this case, both LoRA and Houlsby adapter can achieve more than 88\% CIDEr score of the baseline.
Moreover, utilizing ViT blocks leads to superior performance in comparison to using a linear projection layer.
This improvement can be attributed to the ViT block's enhanced capability in capturing the specific information of the intermediate visual embeddings relative to the linear projection layer.
Houlsby adapter, in particular, attains a CIDEr score of 86.0, which represents approximately 97.4\% of the baseline. Furthermore, LoRA exhibits the highest performance, achieving a CIDEr score of 88.1, equivalent to roughly 99.8\% of the baseline.
In this experiment, it's evident that the use of visual projection can effectively transfer task-specific features to a certain extent.

\hfill \break

\renewcommand{\arraystretch}{1.5}

\begin{table}[t]
    \centering
    \resizebox{.95\columnwidth}{!}{
    % \setlength{\tabcolsep}{12pt}
    % \begin{tabular}{p{4cm}|p{3cm}p{3cm}p{3cm}}
    \begin{tabular}{c|ccc}
    \hline
    & BLEU-4    & CIDEr & Parameters(\%) \\ \hline                           
    FT (A)       & 20.6 & 88.3 & 100.0 \\
    linear \& text decoder      & 15.7 & 67.3 & 61.66 \\
    linear \& Houlsby           & 17.9 & 81.3 & 1.44 \\
    linaer \& LoRA              & 16.1 & 80.5 & \textbf{0.52} \\
    ViT block \& text decoder   & 16.6 & 77.9 & 62.74 \\
    ViT block \& Houlsby        & \textbf{18.2} & 86.0 & 4.18 \\
    ViT block \& LoRA           & 17.9 & \textbf{88.1} & 3.31 \\
    \hline
    \end{tabular}
    }
    \caption{Performance of text decoder tuning with visual projection layer using parameter-efficient methods.}
\label{tab:2}
\end{table}

\subsection{Entire Model Tuning}
Both types of parameter-efficient tuning methods were implemented simultaneously on the visual encoder and text decoder.
For the text decoder, Houlsby adapter and LoRA were utilized, while EVP was employed for the visual encoder.
Additionally, BitFit is exclusively tested on the entire model.
Given that BitFit's fundamental concept does not involve the introduction of additional modules, but instead focuses on fine-tuning biases, it was deemed pertinent to evaluate its impact across the whole model.
The results are shown in Table \ref{tab:3}.

By utilizing the parameter-efficient tuning methods on the whole model, where the trainable parameters are distributed sparsely throughout the model, we observe that the results are reasonably robust.
The lowest performance achieved in this configuration is a CIDEr score of approximately 79.2, which is about 89.7\% of the baseline.
But the lowest BLEU score is 15.4, which is only 74.8\% of the baseline.
This discrepancy can be attributed to the BLEU metric focusing solely on the n-gram overlap between the generated captions and the reference captions, whereas the CIDEr metric takes into account not only n-grams but also the Term Frequency Inverse Document Frequency (TF-IDF) \cite{robertson2004understanding}, which measures the similarity between the generated captions and the reference captions.
The combinations of EVP with LoRA may capture some implicit information, but may not perform as well in representing explicit details, leading to lower BLEU scores.
However, when using the combination of EVP with Houlsby adapter, the BLEU score could acheieve 18.2, which is 88.3\% of the baseline ans the CIDEr score could reach 85.2, which is 96.5\% of the baseline.

\hfill \break

\renewcommand{\arraystretch}{1.5}

\begin{table}[t]
    \centering
    \resizebox{.95\columnwidth}{!}{
    % \setlength{\tabcolsep}{12pt}
    % \begin{tabular}{p{4cm}|p{3cm}p{3cm}p{3cm}}
    \begin{tabular}{c|ccc}
    \hline
    & BLEU-4    & CIDEr & Parameters(\%) \\ \hline                           
    FT (A)       & 20.6 & 88.3 & 100.0 \\ 
    BitFit (A)      & 17.6 & 80.4 & 0.13 \\
    EVP \& Houlsby  & \textbf{18.2} & \textbf{85.2} & 1.47 \\
    EVP-gs \& Houlsby  & \textbf{18.2} & 84.7 & 1.47 \\
    EVP-fft \& Houlsby  & 18.1 & 84.0 & 1.47 \\
    EVP \& LoRA     & 15.4 & 79.2 & 0.55 \\
    EVP-gs \& LoRA     & 15.4 & 79.2 & 0.55 \\
    EVP-fft \& LoRA     & 15.8 & 80.7 & 0.55 \\
    \hline
    \end{tabular}
    }
    \caption{Overall performance of parameter-efficient tuning methods on entire model.}
\label{tab:3}
\end{table}

\subsection{Discussion}
% 加開頭

From our experimental results, we can observe that the handcrafted features used in EVP do not contribute to the enhancement of the model's performance in our case.
The performance of EVP in this scenario is inferior to the other methods when applied individually to the visual encoder.
However, when EVP is combined with the Houlsby adapter or LoRA, effectively dispersing updatable parameters across the entire model, we observe a substantial improvement in performance.

BitFit, when applied across the entirety of the model, can yield decent results without necessitating the insertion of additional modules.
However, when employed exclusively on a partial component of the model, its effectiveness diminishes.

LoRA, in most cases, can achieve satisfactory results, yet its performance is not on par with that of the Houlsby adapter.
These can be attributed to the fact that LoRA is only applied to the attention module within the Transformer block, whereas the Houlsby adapter is implemented throughout the entirety of the Transformer block.
But when LoRA is applied with a ViT block, it can achieve the highest performance, very close to the baseline.
This may because LoRA in the cross-attention modules directly captures the intermediate visual features during tuning, which are more task-specific, and the ViT block can capture more information than the linear projection layer.

Houlsby adapter, whether used individually on the text decoder, applied with a visual projection layer, or combined with EVP, almost outperforms all other methods.
Houlsby adapter, even when deployed independently, comes close to matching the performance achieved by fine-tuning the entire text decoder.
Moreover, when it is applied with EVP, its performance approximates the effectiveness of fine-tuning the entire model, all while only necessitating updates to a mere 1.47\% of the model's parameters.
This demonstrates its stability and flexibility across various scenarios.

\begin{figure}
    \centerline{\includegraphics[width=\linewidth]{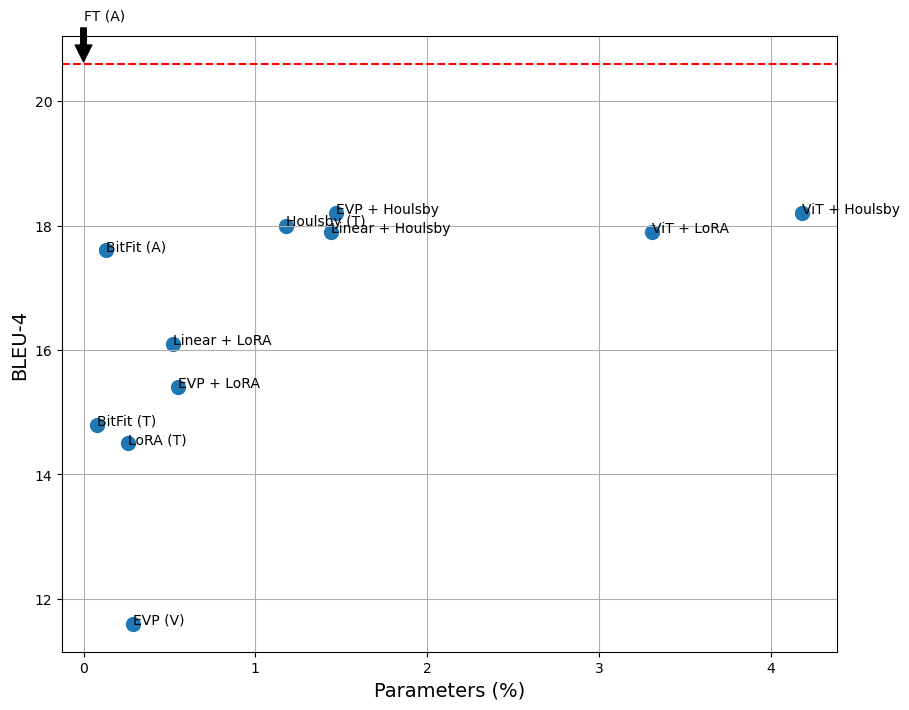}}
    \caption{
        Comparison of the overall BLEU score of parameter-efficient tuning methods.
    }
    \label{fig:5}
\end{figure}

\begin{figure}
    \centerline{\includegraphics[width=\linewidth]{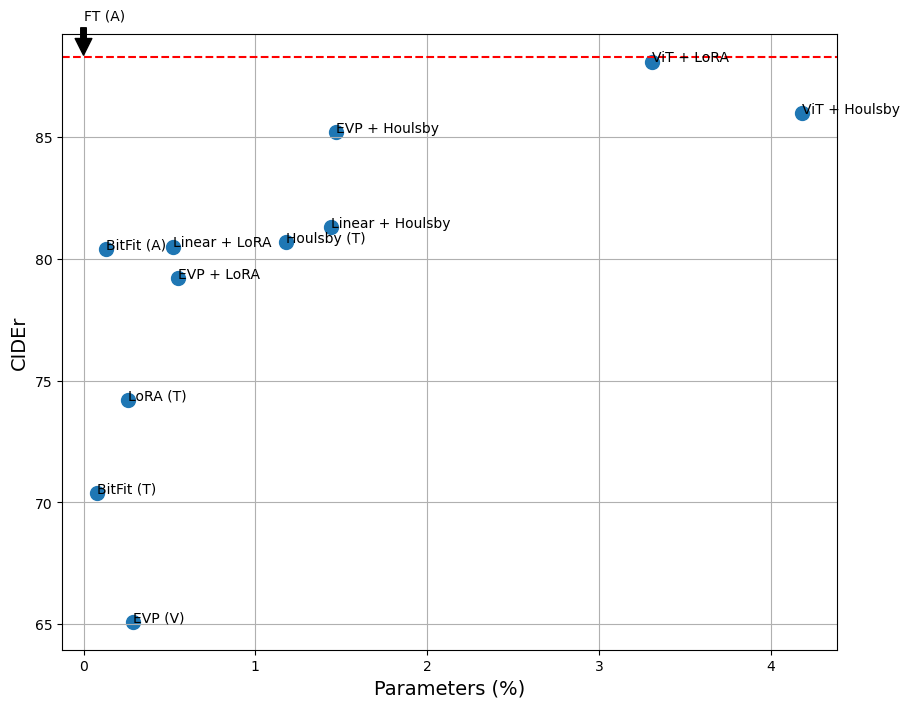}}
    \caption{
        Comparison of the overall CIDEr score of parameter-efficient tuning methods.
    }
    \label{fig:4}
\end{figure}

As demonstrated in Figure \ref{fig:4}, we examine the CIDEr performance of our experimental approaches.
Individually tuning the visual encoder and text decoder using parameter-efficient methods doesn't yield sufficiently satisfactory results.
This could be due to an inadequate number of parameters available to learn task-specific information, or the inability to modify the other component of the model, thereby resulting in model instability.
As depicted in Figure \ref{fig:5}, the BLEU scores of our combined approaches or those using visual projection are all approximately 18, reflecting closely comparable performances.
However, implementing the VL-adapter's model modification with a linear visual projection layer does have some impact, but our application of the ViT block coupled with LoRA—with only 3.31\% of the entire model's parameters—is more effective in capturing intermediate visual features, thus achieving a higher score.
The ViT block, having a larger parameter count than the linear projection layer, could account for its superior performance.
Notably, the ViT block's parameters constitute approximately 3\% of the entire model's parameters, a quantity nearly tenfold that of the EVP's parameters.
The use of EVP with Houlsby adapter, requiring only 1.47\% of the model's parameters, yields a substantial score, reaching 96.5\% of the baseline.
We further delve into comparing the generated captions from these two top-performing methods, as outlined in the Generated Captions section.

% \begin{table*}[t]
% \centering

% \begin{tabular}{l|l|l|l}
% \textbackslash abovecaption &
% \textbackslash abovedisplay &
% \textbackslash addevensidemargin &
% \textbackslash addsidemargin \\
% \textbackslash addtolength &
% \textbackslash baselinestretch &
% \textbackslash belowcaption &
% \textbackslash belowdisplay \\
% \textbackslash break &
% \textbackslash clearpage &
% \textbackslash clip &
% \textbackslash columnsep \\
% \textbackslash float &
% \textbackslash input &
% \textbackslash input &
% \textbackslash linespread \\
% \textbackslash newpage &
% \textbackslash pagebreak &
% \textbackslash renewcommand &
% \textbackslash setlength \\
% \textbackslash text height &
% \textbackslash tiny &
% \textbackslash top margin &
% \textbackslash trim \\
% \textbackslash vskip\{- &
% \textbackslash vspace\{- \\
% \end{tabular}
% %}
% \caption{Commands that must not be used}
% \label{table1}
% \end{table*}

% \begin{table}[t]
% \centering
% %\resizebox{.95\columnwidth}{!}{
% \begin{tabular}{l|l|l|l}
%     authblk & babel & cjk & dvips \\
%     epsf & epsfig & euler & float \\
%     fullpage & geometry & graphics & hyperref \\
%     layout & linespread & lmodern & maltepaper \\
%     navigator & pdfcomment & pgfplots & psfig \\
%     pstricks & t1enc & titlesec & tocbind \\
%     ulem
% \end{tabular}
% \caption{LaTeX style packages that must not be used.}
% \label{table2}
% \end{table}

% \appendix
\section{Generated Captions}
\label{appendix:1}

\nobibliography*
In this section, we present a collection of example captions generated by our model, compared with human labels and the outputs from the Screen2Words paper's model in the context of screenshot captioning tasks.
These examples further elucidate our model's capabilities in generating captions, extending beyond the quantitative language scores covered in the main sections of the paper.
These selections have been curated to showcase the diversity and quality of captions generated across varied scenarios and conditions.
We also include a few examples where our model fails to generate captions that are as accurate as the human labels, and discuss the possible reasons for these failures.
The examples provided here encompass captions generated by our implementation of EVP with the Houlsby adapter, LoRA paired with a ViT block visual projection, and the fine-tuning of the entire model.
For a comparative analysis, we also include the generated captions from the model implemented in the Screen2Words paper.
It's worth noting that the model from the Screen2Words paper is a multi-modal model that uses not only images but also the text and layout information present in the screenshots as inputs.

\begin{figure}
    \centerline{\includegraphics[width=\linewidth]{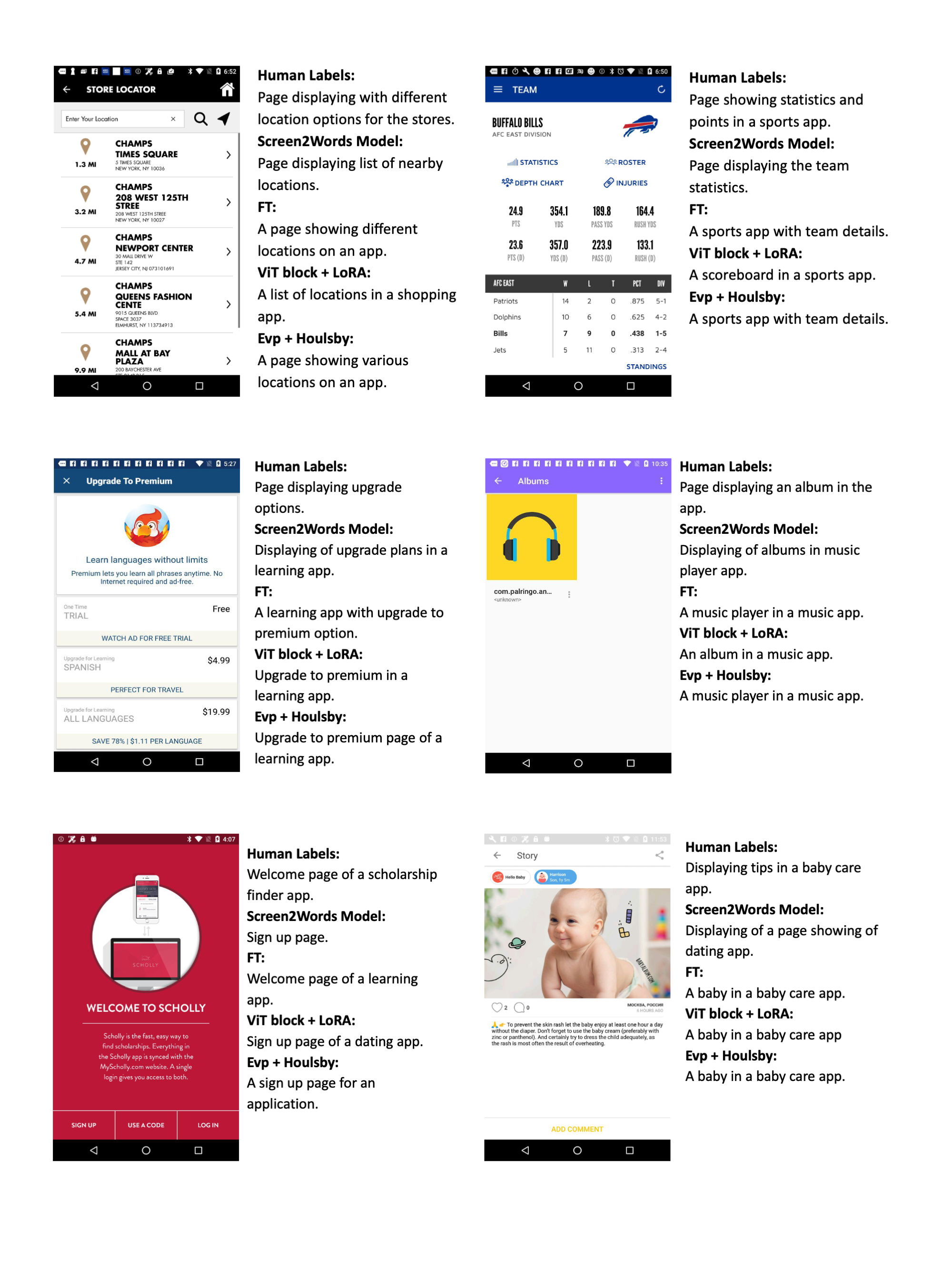}}
    \caption{Example of the generated captions}
    \label{fig:3}
\end{figure}

As depicted in Figure \ref{fig:3}, our models are able to capture the global information but struggle to retrieve the detailed information, as evidenced in the two top examples.
as evidenced in the two top examples.
The failures observed could be attributed to the abundance of text on the screenshot and the fact that we solely use the image as input.
In the middle-left example, our two adapter-based models exhibit a comprehensive understanding of the context as fine-tuning the.
In the middle-right example, the caption generated by LoRA with the ViT block visual projection aligns more closely with the human label than the captions produced by either our implementation of EVP with the Houlsby adapter or by fine-tuning the entire model.
Interestingly, the latter two models mistakenly interpret the album as a music player.
This disfference could potentially be attributed to the LoRA's insertion into the attention module, enabling the model to focus more on semantic features.
In the bottom-left example, our models grasp half of the context—they can identify it as a login page but fail to recognize which app it is.
This may be attributed to the simplicity of the login page, which consists of only a few elements in the page.
Additionally, the models may struggle to comprehend the text within the page.
In the bottom-right example, our models manage to discern that the app is a baby care app, but do not pick up on the text information displayed on the screen.
Notably, the caption describes a baby in the app, which may be due to the pre-training of the model on diverse datasets, including classification datasets.
This pre-training may predispose the model to concentrate on objects present in the image.

\section{Conclusion}

This study explores the effectiveness of various parameter-efficient tuning strategies, originally designed for both visual and language tasks, within the context of screenshot captioning and conduct various combinations of the methods to identify the most effective combination for the BLIP Caption model.
Through our experiments, we find that using LoRA with a ViT block as a visual projection achieved the highest CIDEr performance, reaching about 99.8\% of the performance achieved by fully fine-tuning the model, with only 3.31\% of the model's parameters needing tuning.
Similarly, the combination of EVP on the visual encoder and the Houlsby adapter on the text decoder reached about 96.5\% of the performance of full fine-tuning, using only 1.47\% of the model's parameters.
This underscores that both EVP with Houlsby adapter and LoRA with ViT block are viable choices, depending on the parameter constraints.
It also demonstrates the efficacy of parameter-efficient tuning strategies in enhancing screenshot captioning performance.
These findings serve as an important benchmark for future research in this field.
Moreover, the generated captions by these models for the test set could be a valuable resource for further studies related to caption generation.

\section{Acknowledgments}

This project is funded by Google's grant to National Taiwan University numbered "Google-NTU-112-6-00049".

\bibliography{aaai24}

\begin{thebibliography}{52}
\providecommand{\natexlab}[1]{#1}

\bibitem[{Agrawal et~al.(2019)Agrawal, Desai, Wang, Chen, Jain, Johnson, Batra,
  Parikh, Lee, and Anderson}]{agrawal2019nocaps}
Agrawal, H.; Desai, K.; Wang, Y.; Chen, X.; Jain, R.; Johnson, M.; Batra, D.;
  Parikh, D.; Lee, S.; and Anderson, P. 2019.
\newblock Nocaps: Novel object captioning at scale.
\newblock In \emph{Proceedings of the IEEE/CVF international conference on
  computer vision}, 8948--8957.

\bibitem[{Anderson et~al.(2018)Anderson, He, Buehler, Teney, Johnson, Gould,
  and Zhang}]{anderson2018bottom}
Anderson, P.; He, X.; Buehler, C.; Teney, D.; Johnson, M.; Gould, S.; and
  Zhang, L. 2018.
\newblock Bottom-up and top-down attention for image captioning and visual
  question answering.
\newblock In \emph{Proceedings of the IEEE conference on computer vision and
  pattern recognition}, 6077--6086.

\bibitem[{Bapna, Arivazhagan, and Firat(2019)}]{bapna2019simple}
Bapna, A.; Arivazhagan, N.; and Firat, O. 2019.
\newblock Simple, scalable adaptation for neural machine translation.
\newblock \emph{arXiv preprint arXiv:1909.08478}.

\bibitem[{Changpinyo et~al.(2021)Changpinyo, Sharma, Ding, and
  Soricut}]{changpinyo2021conceptual}
Changpinyo, S.; Sharma, P.; Ding, N.; and Soricut, R. 2021.
\newblock Conceptual 12m: Pushing web-scale image-text pre-training to
  recognize long-tail visual concepts.
\newblock In \emph{Proceedings of the IEEE/CVF Conference on Computer Vision
  and Pattern Recognition}, 3558--3568.

\bibitem[{Chen and Dolan(2011)}]{chen-dolan-2011-collecting}
Chen, D.; and Dolan, W. 2011.
\newblock Collecting Highly Parallel Data for Paraphrase Evaluation.
\newblock In \emph{Proceedings of the 49th Annual Meeting of the Association
  for Computational Linguistics: Human Language Technologies}, 190--200.
  Portland, Oregon, USA: Association for Computational Linguistics.

\bibitem[{Chen et~al.(2023{\natexlab{a}})Chen, Zhu, Ding, Cao, Wang, Li, Sun,
  Mao, and Zang}]{chen2023sam}
Chen, T.; Zhu, L.; Ding, C.; Cao, R.; Wang, Y.; Li, Z.; Sun, L.; Mao, P.; and
  Zang, Y. 2023{\natexlab{a}}.
\newblock SAM Fails to Segment Anything? -- SAM-Adapter: Adapting SAM in
  Underperformed Scenes: Camouflage, Shadow, Medical Image Segmentation, and
  More.
\newblock arXiv:2304.09148.

\bibitem[{Chen et~al.(2023{\natexlab{b}})Chen, Fu, Liu, Li, and
  Lee}]{chen2023exploring}
Chen, Z.-C.; Fu, C.-L.; Liu, C.-Y.; Li, S.-W.~D.; and Lee, H.-y.
  2023{\natexlab{b}}.
\newblock Exploring efficient-tuning methods in self-supervised speech models.
\newblock In \emph{2022 IEEE Spoken Language Technology Workshop (SLT)},
  1120--1127. IEEE.

\bibitem[{Deka et~al.(2017)Deka, Huang, Franzen, Hibschman, Afergan, Li,
  Nichols, and Kumar}]{deka2017rico}
Deka, B.; Huang, Z.; Franzen, C.; Hibschman, J.; Afergan, D.; Li, Y.; Nichols,
  J.; and Kumar, R. 2017.
\newblock Rico: A mobile app dataset for building data-driven design
  applications.
\newblock In \emph{Proceedings of the 30th annual ACM symposium on user
  interface software and technology}, 845--854.

\bibitem[{Deng et~al.(2009)Deng, Dong, Socher, Li, Li, and Fei-Fei}]{5206848}
Deng, J.; Dong, W.; Socher, R.; Li, L.-J.; Li, K.; and Fei-Fei, L. 2009.
\newblock ImageNet: A large-scale hierarchical image database.
\newblock In \emph{2009 IEEE Conference on Computer Vision and Pattern
  Recognition}, 248--255.

\bibitem[{Devlin et~al.(2019)Devlin, Chang, Lee, and
  Toutanova}]{devlin-etal-2019-bert}
Devlin, J.; Chang, M.-W.; Lee, K.; and Toutanova, K. 2019.
\newblock {BERT}: Pre-training of Deep Bidirectional Transformers for Language
  Understanding.
\newblock In \emph{Proceedings of the 2019 Conference of the North {A}merican
  Chapter of the Association for Computational Linguistics: Human Language
  Technologies, Volume 1 (Long and Short Papers)}, 4171--4186. Minneapolis,
  Minnesota: Association for Computational Linguistics.

\bibitem[{Dosovitskiy et~al.(2021)Dosovitskiy, Beyer, Kolesnikov, Weissenborn,
  Zhai, Unterthiner, Dehghani, Minderer, Heigold, Gelly, Uszkoreit, and
  Houlsby}]{dosovitskiy2021an}
Dosovitskiy, A.; Beyer, L.; Kolesnikov, A.; Weissenborn, D.; Zhai, X.;
  Unterthiner, T.; Dehghani, M.; Minderer, M.; Heigold, G.; Gelly, S.;
  Uszkoreit, J.; and Houlsby, N. 2021.
\newblock An Image is Worth 16x16 Words: Transformers for Image Recognition at
  Scale.
\newblock In \emph{International Conference on Learning Representations}.

\bibitem[{Ermis et~al.(2022)Ermis, Zappella, Wistuba, Rawal, and
  Archambeau}]{Ermis_2022_CVPR}
Ermis, B.; Zappella, G.; Wistuba, M.; Rawal, A.; and Archambeau, C. 2022.
\newblock Continual Learning With Transformers for Image Classification.
\newblock In \emph{Proceedings of the IEEE/CVF Conference on Computer Vision
  and Pattern Recognition (CVPR) Workshops}, 3774--3781.

\bibitem[{Guo, Rush, and Kim(2020)}]{guo2020parameter}
Guo, D.; Rush, A.~M.; and Kim, Y. 2020.
\newblock Parameter-efficient transfer learning with diff pruning.
\newblock \emph{arXiv preprint arXiv:2012.07463}.

\bibitem[{He et~al.(2021)He, Zhou, Ma, Berg{-}Kirkpatrick, and
  Neubig}]{DBLP:journals/corr/abs-2110-04366}
He, J.; Zhou, C.; Ma, X.; Berg{-}Kirkpatrick, T.; and Neubig, G. 2021.
\newblock Towards a Unified View of Parameter-Efficient Transfer Learning.
\newblock \emph{CoRR}, abs/2110.04366.

\bibitem[{Hochreiter and Schmidhuber(1997)}]{hochreiter1997long}
Hochreiter, S.; and Schmidhuber, J. 1997.
\newblock Long short-term memory.
\newblock \emph{Neural computation}, 9(8): 1735--1780.

\bibitem[{Houlsby et~al.(2019)Houlsby, Giurgiu, Jastrzebski, Morrone,
  De~Laroussilhe, Gesmundo, Attariyan, and Gelly}]{houlsby2019parameter}
Houlsby, N.; Giurgiu, A.; Jastrzebski, S.; Morrone, B.; De~Laroussilhe, Q.;
  Gesmundo, A.; Attariyan, M.; and Gelly, S. 2019.
\newblock Parameter-efficient transfer learning for NLP.
\newblock In \emph{International Conference on Machine Learning}, 2790--2799.
  PMLR.

\bibitem[{Hu et~al.(2021)Hu, Shen, Wallis, Allen-Zhu, Li, Wang, Wang, and
  Chen}]{hu2021lora}
Hu, E.~J.; Shen, Y.; Wallis, P.; Allen-Zhu, Z.; Li, Y.; Wang, S.; Wang, L.; and
  Chen, W. 2021.
\newblock Lora: Low-rank adaptation of large language models.
\newblock \emph{arXiv preprint arXiv:2106.09685}.

\bibitem[{Karimi~Mahabadi, Henderson, and Ruder(2021)}]{karimi2021compacter}
Karimi~Mahabadi, R.; Henderson, J.; and Ruder, S. 2021.
\newblock Compacter: Efficient low-rank hypercomplex adapter layers.
\newblock \emph{Advances in Neural Information Processing Systems}, 34:
  1022--1035.

\bibitem[{Krishna et~al.(2017)Krishna, Zhu, Groth, Johnson, Hata, Kravitz,
  Chen, Kalantidis, Li, Shamma et~al.}]{krishna2017visual}
Krishna, R.; Zhu, Y.; Groth, O.; Johnson, J.; Hata, K.; Kravitz, J.; Chen, S.;
  Kalantidis, Y.; Li, L.-J.; Shamma, D.~A.; et~al. 2017.
\newblock Visual genome: Connecting language and vision using crowdsourced
  dense image annotations.
\newblock \emph{International journal of computer vision}, 123: 32--73.

\bibitem[{Krizhevsky, Sutskever, and Hinton(2012)}]{krizhevsky2012imagenet}
Krizhevsky, A.; Sutskever, I.; and Hinton, G.~E. 2012.
\newblock Imagenet classification with deep convolutional neural networks.
\newblock \emph{Advances in neural information processing systems}, 25.

\bibitem[{Li et~al.(2022{\natexlab{a}})Li, Li, Le, Wang, Savarese, and
  Hoi}]{li2022lavis}
Li, D.; Li, J.; Le, H.; Wang, G.; Savarese, S.; and Hoi, S. C.~H.
  2022{\natexlab{a}}.
\newblock LAVIS: A Library for Language-Vision Intelligence.
\newblock arXiv:2209.09019.

\bibitem[{Li et~al.(2023)Li, Li, Savarese, and Hoi}]{li2023blip}
Li, J.; Li, D.; Savarese, S.; and Hoi, S. 2023.
\newblock Blip-2: Bootstrapping language-image pre-training with frozen image
  encoders and large language models.
\newblock \emph{arXiv preprint arXiv:2301.12597}.

\bibitem[{Li et~al.(2022{\natexlab{b}})Li, Li, Xiong, and Hoi}]{li2022blip}
Li, J.; Li, D.; Xiong, C.; and Hoi, S. 2022{\natexlab{b}}.
\newblock Blip: Bootstrapping language-image pre-training for unified
  vision-language understanding and generation.
\newblock In \emph{International Conference on Machine Learning}, 12888--12900.
  PMLR.

\bibitem[{Li et~al.(2021)Li, Selvaraju, Gotmare, Joty, Xiong, and
  Hoi}]{li2021align}
Li, J.; Selvaraju, R.; Gotmare, A.; Joty, S.; Xiong, C.; and Hoi, S. C.~H.
  2021.
\newblock Align before fuse: Vision and language representation learning with
  momentum distillation.
\newblock \emph{Advances in neural information processing systems}, 34:
  9694--9705.

\bibitem[{Li and Liang(2021)}]{li2021prefix}
Li, X.~L.; and Liang, P. 2021.
\newblock Prefix-tuning: Optimizing continuous prompts for generation.
\newblock \emph{arXiv preprint arXiv:2101.00190}.

\bibitem[{Lin et~al.(2014)Lin, Maire, Belongie, Hays, Perona, Ramanan,
  Doll{\'a}r, and Zitnick}]{10.1007/978-3-319-10602-1_48}
Lin, T.-Y.; Maire, M.; Belongie, S.; Hays, J.; Perona, P.; Ramanan, D.;
  Doll{\'a}r, P.; and Zitnick, C.~L. 2014.
\newblock Microsoft COCO: Common Objects in Context.
\newblock In Fleet, D.; Pajdla, T.; Schiele, B.; and Tuytelaars, T., eds.,
  \emph{Computer Vision -- ECCV 2014}, 740--755. Cham: Springer International
  Publishing.
\newblock ISBN 978-3-319-10602-1.

\bibitem[{Liu et~al.(2023)Liu, Shen, Pun, and Cun}]{liu2023explicit}
Liu, W.; Shen, X.; Pun, C.-M.; and Cun, X. 2023.
\newblock Explicit visual prompting for low-level structure segmentations.
\newblock In \emph{Proceedings of the IEEE/CVF Conference on Computer Vision
  and Pattern Recognition}, 19434--19445.

\bibitem[{Liu et~al.(2021)Liu, Lin, Cao, Hu, Wei, Zhang, Lin, and
  Guo}]{liu2021swin}
Liu, Z.; Lin, Y.; Cao, Y.; Hu, H.; Wei, Y.; Zhang, Z.; Lin, S.; and Guo, B.
  2021.
\newblock Swin transformer: Hierarchical vision transformer using shifted
  windows.
\newblock In \emph{Proceedings of the IEEE/CVF international conference on
  computer vision}, 10012--10022.

\bibitem[{Ordonez, Kulkarni, and Berg(2011)}]{ordonez2011im2text}
Ordonez, V.; Kulkarni, G.; and Berg, T. 2011.
\newblock Im2text: Describing images using 1 million captioned photographs.
\newblock \emph{Advances in neural information processing systems}, 24.

\bibitem[{Pan et~al.(2022)Pan, Lin, Zhu, Shao, and Li}]{pan2022st}
Pan, J.; Lin, Z.; Zhu, X.; Shao, J.; and Li, H. 2022.
\newblock ST-Adapter: Parameter-Efficient Image-to-Video Transfer Learning for
  Action Recognition.
\newblock \emph{arXiv preprint arXiv:2206.13559}.

\bibitem[{Papineni et~al.(2002)Papineni, Roukos, Ward, and
  Zhu}]{papineni2002bleu}
Papineni, K.; Roukos, S.; Ward, T.; and Zhu, W.-J. 2002.
\newblock Bleu: a method for automatic evaluation of machine translation.
\newblock In \emph{Proceedings of the 40th annual meeting of the Association
  for Computational Linguistics}, 311--318.

\bibitem[{Paszke et~al.(2019)Paszke, Gross, Massa, Lerer, Bradbury, Chanan,
  Killeen, Lin, Gimelshein, Antiga et~al.}]{paszke2019pytorch}
Paszke, A.; Gross, S.; Massa, F.; Lerer, A.; Bradbury, J.; Chanan, G.; Killeen,
  T.; Lin, Z.; Gimelshein, N.; Antiga, L.; et~al. 2019.
\newblock Pytorch: An imperative style, high-performance deep learning library.
\newblock \emph{Advances in neural information processing systems}, 32.

\bibitem[{Pfeiffer et~al.(2020)Pfeiffer, R{\"u}ckl{\'e}, Poth, Kamath,
  Vuli{\'c}, Ruder, Cho, and Gurevych}]{pfeiffer2020AdapterHub}
Pfeiffer, J.; R{\"u}ckl{\'e}, A.; Poth, C.; Kamath, A.; Vuli{\'c}, I.; Ruder,
  S.; Cho, K.; and Gurevych, I. 2020.
\newblock AdapterHub: A Framework for Adapting Transformers.
\newblock In \emph{Proceedings of the 2020 Conference on Empirical Methods in
  Natural Language Processing: System Demonstrations}, 46--54.

\bibitem[{Radford et~al.(2021)Radford, Kim, Hallacy, Ramesh, Goh, Agarwal,
  Sastry, Askell, Mishkin, Clark et~al.}]{radford2021learning}
Radford, A.; Kim, J.~W.; Hallacy, C.; Ramesh, A.; Goh, G.; Agarwal, S.; Sastry,
  G.; Askell, A.; Mishkin, P.; Clark, J.; et~al. 2021.
\newblock Learning transferable visual models from natural language
  supervision.
\newblock In \emph{International conference on machine learning}, 8748--8763.
  PMLR.

\bibitem[{Raffel et~al.(2020)Raffel, Shazeer, Roberts, Lee, Narang, Matena,
  Zhou, Li, and Liu}]{raffel2020exploring}
Raffel, C.; Shazeer, N.; Roberts, A.; Lee, K.; Narang, S.; Matena, M.; Zhou,
  Y.; Li, W.; and Liu, P.~J. 2020.
\newblock Exploring the limits of transfer learning with a unified text-to-text
  transformer.
\newblock \emph{The Journal of Machine Learning Research}, 21(1): 5485--5551.

\bibitem[{Ren et~al.(2015)Ren, He, Girshick, and Sun}]{ren2015faster}
Ren, S.; He, K.; Girshick, R.; and Sun, J. 2015.
\newblock Faster r-cnn: Towards real-time object detection with region proposal
  networks.
\newblock \emph{Advances in neural information processing systems}, 28.

\bibitem[{Robertson(2004)}]{robertson2004understanding}
Robertson, S. 2004.
\newblock Understanding inverse document frequency: on theoretical arguments
  for IDF.
\newblock \emph{Journal of documentation}, 60(5): 503--520.

\bibitem[{Schuhmann et~al.(2021)Schuhmann, Vencu, Beaumont, Kaczmarczyk,
  Mullis, Katta, Coombes, Jitsev, and Komatsuzaki}]{schuhmann2021laion}
Schuhmann, C.; Vencu, R.; Beaumont, R.; Kaczmarczyk, R.; Mullis, C.; Katta, A.;
  Coombes, T.; Jitsev, J.; and Komatsuzaki, A. 2021.
\newblock Laion-400m: Open dataset of clip-filtered 400 million image-text
  pairs.
\newblock \emph{arXiv preprint arXiv:2111.02114}.

\bibitem[{Sharma et~al.(2018)Sharma, Ding, Goodman, and
  Soricut}]{sharma2018conceptual}
Sharma, P.; Ding, N.; Goodman, S.; and Soricut, R. 2018.
\newblock Conceptual captions: A cleaned, hypernymed, image alt-text dataset
  for automatic image captioning.
\newblock In \emph{Proceedings of the 56th Annual Meeting of the Association
  for Computational Linguistics (Volume 1: Long Papers)}, 2556--2565.

\bibitem[{Subramanian et~al.(2020)Subramanian, Wang, Mehta, Bogin, van Zuylen,
  Parasa, Singh, Gardner, and Hajishirzi}]{subramanian2020medicat}
Subramanian, S.; Wang, L.~L.; Mehta, S.; Bogin, B.; van Zuylen, M.; Parasa, S.;
  Singh, S.; Gardner, M.; and Hajishirzi, H. 2020.
\newblock Medicat: A dataset of medical images, captions, and textual
  references.
\newblock \emph{arXiv preprint arXiv:2010.06000}.

\bibitem[{Sung, Cho, and Bansal(2022)}]{sung2022vl}
Sung, Y.-L.; Cho, J.; and Bansal, M. 2022.
\newblock Vl-adapter: Parameter-efficient transfer learning for
  vision-and-language tasks.
\newblock In \emph{Proceedings of the IEEE/CVF Conference on Computer Vision
  and Pattern Recognition}, 5227--5237.

\bibitem[{Vaswani et~al.(2017)Vaswani, Shazeer, Parmar, Uszkoreit, Jones,
  Gomez, Kaiser, and Polosukhin}]{NIPS2017_3f5ee243}
Vaswani, A.; Shazeer, N.; Parmar, N.; Uszkoreit, J.; Jones, L.; Gomez, A.~N.;
  Kaiser, L.~u.; and Polosukhin, I. 2017.
\newblock Attention is All you Need.
\newblock In Guyon, I.; Luxburg, U.~V.; Bengio, S.; Wallach, H.; Fergus, R.;
  Vishwanathan, S.; and Garnett, R., eds., \emph{Advances in Neural Information
  Processing Systems}, volume~30. Curran Associates, Inc.

\bibitem[{Vedantam, Lawrence~Zitnick, and Parikh(2015)}]{vedantam2015cider}
Vedantam, R.; Lawrence~Zitnick, C.; and Parikh, D. 2015.
\newblock Cider: Consensus-based image description evaluation.
\newblock In \emph{Proceedings of the IEEE conference on computer vision and
  pattern recognition}, 4566--4575.

\bibitem[{Vinyals et~al.(2015)Vinyals, Toshev, Bengio, and
  Erhan}]{vinyals2015show}
Vinyals, O.; Toshev, A.; Bengio, S.; and Erhan, D. 2015.
\newblock Show and tell: A neural image caption generator.
\newblock In \emph{Proceedings of the IEEE conference on computer vision and
  pattern recognition}, 3156--3164.

\bibitem[{Wang et~al.(2021)Wang, Li, Zhou, Chen, Grossman, and
  Li}]{wang2021screen2words}
Wang, B.; Li, G.; Zhou, X.; Chen, Z.; Grossman, T.; and Li, Y. 2021.
\newblock Screen2words: Automatic mobile UI summarization with multimodal
  learning.
\newblock In \emph{The 34th Annual ACM Symposium on User Interface Software and
  Technology}, 498--510.

\bibitem[{Xu et~al.(2016)Xu, Mei, Yao, and Rui}]{xu2016msr}
Xu, J.; Mei, T.; Yao, T.; and Rui, Y. 2016.
\newblock Msr-vtt: A large video description dataset for bridging video and
  language.
\newblock In \emph{Proceedings of the IEEE conference on computer vision and
  pattern recognition}, 5288--5296.

\bibitem[{Yang et~al.(2022)Yang, Qiao, Yu, Yuan, Zhu, Yuille, Adam, and
  Chen}]{yang2022moat}
Yang, C.; Qiao, S.; Yu, Q.; Yuan, X.; Zhu, Y.; Yuille, A.; Adam, H.; and Chen,
  L.-C. 2022.
\newblock Moat: Alternating mobile convolution and attention brings strong
  vision models.
\newblock \emph{arXiv preprint arXiv:2210.01820}.

\bibitem[{Yi-Lin~Sung(2022)}]{sung2022vladapter}
Yi-Lin~Sung, M.~B., Jaemin~Cho. 2022.
\newblock VL-Adapter: Parameter-Efficient Transfer Learning for
  Vision-and-Language Tasks.
\newblock In \emph{CVPR}.

\bibitem[{Zaken, Ravfogel, and Goldberg(2021)}]{zaken2021bitfit}
Zaken, E.~B.; Ravfogel, S.; and Goldberg, Y. 2021.
\newblock Bitfit: Simple parameter-efficient fine-tuning for transformer-based
  masked language-models.
\newblock \emph{arXiv preprint arXiv:2106.10199}.

\bibitem[{Zhou et~al.(2022)Zhou, Yang, Loy, and Liu}]{zhou2022learning}
Zhou, K.; Yang, J.; Loy, C.~C.; and Liu, Z. 2022.
\newblock Learning to prompt for vision-language models.
\newblock \emph{International Journal of Computer Vision}, 130(9): 2337--2348.

\bibitem[{Zhou et~al.(2019)Zhou, Palangi, Zhang, Hu, Corso, and
  Gao}]{DBLP:journals/corr/abs-1909-11059}
Zhou, L.; Palangi, H.; Zhang, L.; Hu, H.; Corso, J.~J.; and Gao, J. 2019.
\newblock Unified Vision-Language Pre-Training for Image Captioning and {VQA}.
\newblock \emph{CoRR}, abs/1909.11059.

\bibitem[{Zia, {Mohsin Riaz}, and Ghafoor(2022)}]{ZIA2022102741}
Zia, U.; {Mohsin Riaz}, M.; and Ghafoor, A. 2022.
\newblock Transforming remote sensing images to textual descriptions.
\newblock \emph{International Journal of Applied Earth Observation and
  Geoinformation}, 108: 102741.

\end{thebibliography}

\end{document}